# Sketch Layer Separation in Multi-Spectral Historical Document Images

AmirAbbas Davari
Friedrich-Alexander-University Erlangen-Nuremberg, Erlangen, Germany
Computer Science Department, Pattern Recognition Lab
amir.davari@fau.de

Armin Häberle
Bibliotheca Hertziana - Max-Planck-Institute for Art History, Rome, Italy
haeberle@biblhertz.it

Vincent Christlein
Friedrich-Alexander-University Erlangen-Nuremberg, Erlangen, Germany
Computer Science Department, Pattern Recognition Lab
vincent.christlein@cs.fau.de

Andreas Maier
Friedrich-Alexander-University Erlangen-Nuremberg, Erlangen, Germany
Computer Science Department, Pattern Recognition Lab
andreas.maier@fau.de

Christian Riess
Friedrich-Alexander-University Erlangen-Nuremberg, Erlangen, Germany
Computer Science Department, IT Security Infrastructures
christian.riess@fau.de

**Abstract**

High-resolution imaging has delivered new prospects for detecting the material composition and structure of cultural treasures. Despite the various techniques for analysis, a significant diagnostic gap remained in the range of available research capabilities for works on paper.
Old master drawings were mostly composed in a multi-step manner with various materials. This resulted in the overlapping of different layers which made the subjacent strata difficult to differentiate. The separation of stratified layers using imaging methods could provide insights into the artistic work processes and help answer questions about the object, its attribution, or in identifying forgeries. The pattern recognition procedure was tested with mock replicas to achieve the separation and the capability of displaying concealed red chalk under ink. In contrast to RGB-sensor based imaging, the multi- or hyperspectral technology allows accurate layer separation by recording the characteristic signatures of the material's reflectance. The risk of damage to the artworks as a result of the examination can be reduced by using combinations of defined spectra for lightning and image capturing. By guaranteeing the maximum level of readability, our results suggest that the technique can be applied to a broader range of objects and assist in diagnostic research into cultural treasures in the future.

**Introduction**

The considerable progress in computer technologies and informatics to date has only recently made it possible to build research links between these fields and art history. Despite the long tradition of integrating conventional radiation-based diagnostics into conservation science for the evaluation of artworks, brand-new digital methods for recognizing patterns and processing large data sets have not been incorporated into the standard approach. These emerging technical systems yield completely new tools, and generate further valuable research questions (Beck 1995, p. 622; Stork 2009). The need to find non-destructive methods of examination that could provide much more accurate criteria for material analysis, dating, attributing, and identifying forgeries has determined the type of examinations requested by art historians (Beck 1995, p. 622; cf. Young 1965 and Young 1975).

Furthermore, the use of scientific and multispectral imaging techniques in the investigation of art objects was long confined to the analysis of paintings (e.g. Young 1965 and 1975; Mairinger 2003; Capellini et al. 2003, Pelagotti et al. 2008; Barni et al. 2008; Fairies 2009, Cotte 2010, Cosentino 2014; Pronti et al. 2015). In the research into old master drawings, interest in new methods and approaches came mostly from restorers. Partially due to the limited number of well known drawing materials, as well as a reliance on essential but dated publications such as Meder's fundamental treatise on Old Master drawings, connoisseurship has long taken for granted the number of diverse and complex methods of artistic execution (Meder 1919; Meder & Ames 1975; Häberle 2016a).

From the medieval period until the end of the nineteenth century, drawings were created mostly in a multi-step manner with various materials and techniques. For example, in the corpus of drawings by Nicolas Poussin (cf. Rosenberg & Prat 1994), approximately 75% are multi-material drawings. In the final drawing, multiple layers of different substrates overlap, which means the first layers applied by the artist are often unidentifiable and undetected visually by the unaided eye (Figure 1).

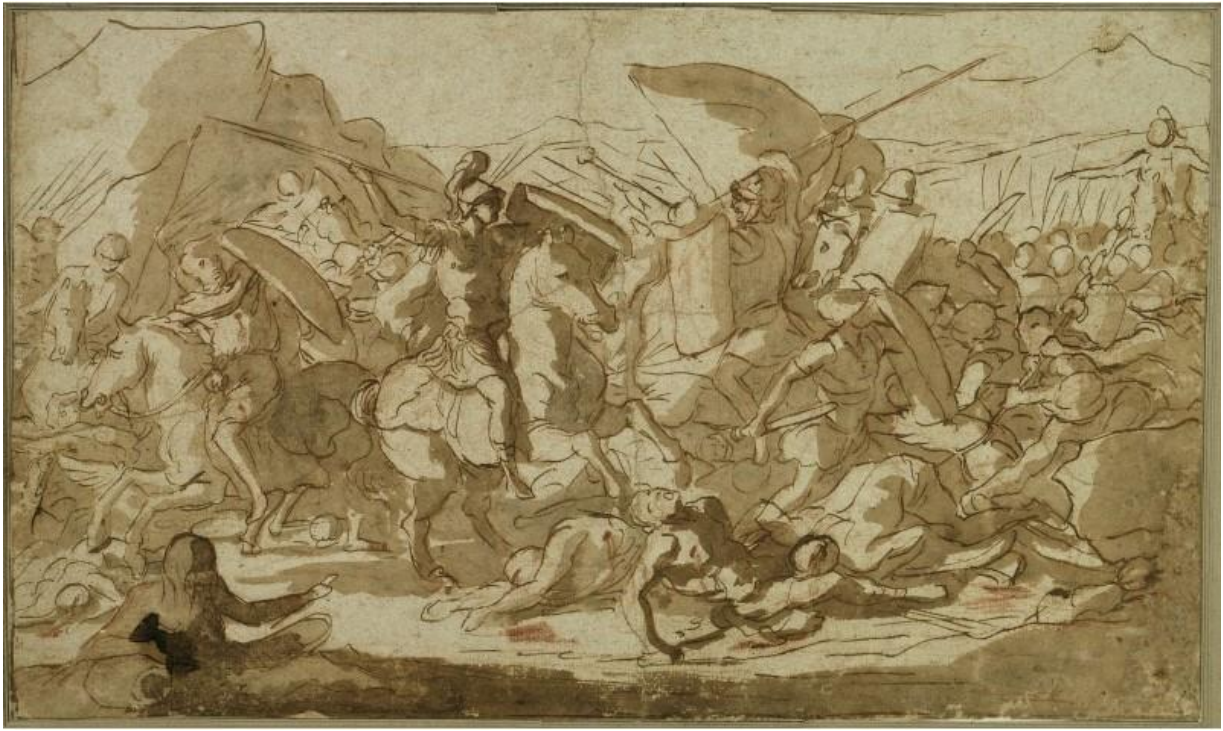

Figure 1: Nicolas Poussin (*1594 - †1665): "The Victory of Joshua over the Amorites", ~1630*
Red chalk, pen and brown ink, black chalk, brown wash, reworked with red chalk, on paper,
246 x 410 mm, Kupferstichkabinett Berlin, KdZ 20887
© Photo: Kupferstichkabinett der Staatlichen Museen zu Berlin - Preußischer Kulturbesitz,
Photographer: Jörg P. Anders

The different substrates, however, illustrate the various steps involved in the genesis of the artwork. Over the centuries, we find numerous techniques of laying down a preliminary drawing in dry media (mostly chalks or graphite) followed by a subsequent wet layer (pen or brush with ink). Exceptions naturally exist and, in these instances, the artist returned to a dry medium after laying down an ink wash. Fig. 1 is a clear example: The artist marked the bleeding wounds of the warriors with red chalk at a later time. (*For iconography and work process analysis cf. Häberle 2016b pp. 51-59, 79-81, 110, fig. 1). For this reason, the preliminary dry drawing could be considered to be mostly finished by the time the artist changed to the wet technique. Thus, the first layer of the drawing represents the starting point of the creative process and represents a crucial point of investigation.

The separation of drawing layers and the visualisation of these strata by means of imaging and image processing promises a direct insight into the artistic process that was previously unavailable. The use of this technology allows for a better understanding of both the genesis of an artwork and the artist's creative interventions. This technique provides art historians with the visualization of each step of the drawing process, translating the results of the technical analysis into a readable image. By understanding the stages of the artistic process, it becomes possible to answer crucial questions about both the object (its formation and evolution) and the artist's working technique.

It is noteworthy that the enhanced sensitivity and increasingly detailed results of scientific methods in the last decade brought striking findings from new examinations of manuscripts and palimpsests (Knox 2008; Walvoord et al. 2008). These techniques delivered significant insights in the analysis of drawings. Using cutting-edge, high-resolution image acquisition and data processing have provided pertinent information about the object's material structure (Knox 2008; Walvoord et al. 2008; Brahms et al. 2008). During the last ten years, larger groups of Old Master drawings were systematically subjected to an extended spectrum of imaging analysis (Ambers et al. 2010; Brahms et al. 2008). Although there are today many specialized examination methods, research into Old Master drawings still suffers from major diagnostic deficiencies:

a) The reliable high-resolution visualization of overlapping material layers over the whole spectrum of drawing media and the ability to image small traces of graphite.
b) The separation of the representations or writings on the verso that show through the paper and that interfere with the drawing on the recto, which often occur by corrosion caused by iron-gall ink. This is important because the majority of drawings are mounted on paper, so that the verso cannot be inspected.
c) The reconstruction of drawings in the case of erasures, damage and surface abrasions.

Due to the physical properties of red chalk, it remained impossible to visualize a preliminary drawing sufficiently in this media if it had been overlaid with ink. Red chalk is made primarily of natural red clay containing iron oxide and, until the late nineteenth century, it was an especially favoured drawing media (Brachert 2001, pp. 206-207; Eastaugh et al. 2004, pp. 320-321). In the infrared spectrum, for wavelengths above 2000 nm, overlapping layers of ink become transparent and reveal the underlying structures. However, in this range of wavelengths, red chalk shows very similar modes of light reflection as the image carrier itself (i.e. the paper or parchment). As a consequence, this range of wavelengths cannot be used to image overpainted strata of red chalk (cf. Burmester/Renger 1986; Mairinger 2003, pp. (138-46, 146-7). The difficulties of displaying and distinguishing the drawn strata by conventional infrared reflectography (IRR) or with remission-spectroscopy further complicate and obscure the detection and clarity of substrate layers. This can be seen very clearly in the comparative sequence of images from the apocryphal Rembrandt drawings in Munich (visible spectrum versus infrared imaging), published by Burmester and Renger in 1986 (pp. 19-31, Figs. 7a-21b). In this work, we investigate an approach to close this diagnostic gap by applying multispectral analysis and pattern recognition procedures. We field-tested these procedures on drawings that were created to mimic the original work process exactly.

**Assumptions for the development of new imaging procedures**

The development of new technical processes for the examination of artworks – especially for works on paper or parchment – poses a challenge because of the fragility and severe sensitivity of these materials and the colours to all forms of electromagnetic radiation. For almost all paper- and parchment-based art, conservators define three basic defining characteristics (Schmits 2004; Mantler & Klikovits 2004):

1. Light- and radiation-induced damage is cumulative and irreversible;
2. Every lux second or Gray of exposure counts;
3. The objects do not undergo any "recovery" or "regeneration" in total darkness.

As a consequence, analysing drawings or manuscripts results in a certain amount of irreversible, substantial damage (Saunders 2006; Schmits 2004; Mantler & Klikovits 2004). In these cases, art historians, conservators, and scientists balance the benefits of the technical analysis and scientific knowledge against the damage caused by the exposure to harmful light or radiation (Mantler & Klikovits 2004). From an ethical perspective, any analysis conducted before the most recent technical developments and the advent of new procedures is unacceptable s a result of the likely degradation of the object. Serial image acquisitions are considered analogous if the concrete amount of knowledge to be gained cannot be quantified in relation to a single object (cf. Ambers et al. 2010, pp. 7-22).

The most suitable approach to analysing these materials arises from working primarily with samples that are close in material composition and physical properties to the original objects. By doing so, researchers can test the appropriateness of the proposed method and address any problem that occurs during the development process. In the production of these samples, the aesthetic criteria were less relevant than the material composition. A simpler delineation often facilitates the evaluation of the procedure or uncovers additional errors, for example in the case of strokes and hatchings overlaid with ink (Fig. 2).

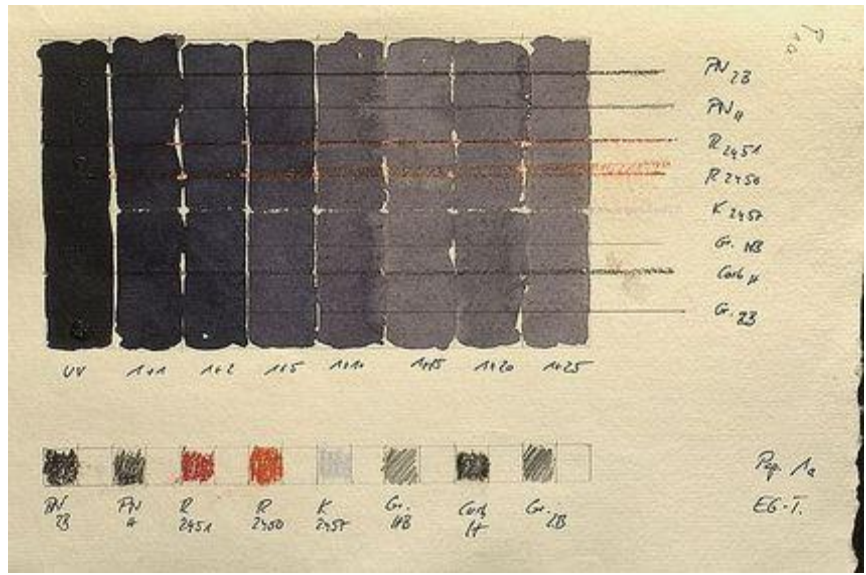

Figure 2: Sample sheet for basic tests.

Strips of iron gall ink (pure ink diluted to 1:25 with water) overlaying strokes of natural black chalk (PN = Pierre noire), red chalk (R), white chalk (K), graphite (G), willow charcoal (Carb). The ancillary letters and numbers mark different grades of hardness or defined colour variants (brownish/reddish), on hand made paper type 1a (= 100% cotton fibers, natural white)

At this point in the project, we image only phantom data. We plan to image old master drawings in our future work, after the efficacy of the proposed approach has been sufficiently proven.

**Experimental Setup**

We created a set of sketches with multiple layers of graphite, chalk and different inks of the same chemical composition that were commonly used in Old Master drawings. After each layer was drawn, the picture was scanned with a book scanner (RGB mode, Zeutschel OS 12000). The step-by-step documentation of the controlled creation process provides knowledge about the drawing layers while also ensuring the accuracy of the method of analysis. Moreover, it allowed us to calibrate the true layer composition for evaluating the layer separation algorithm by subtracting two subsequent scanned images. A sample sketch from this data is shown in Figure 3.

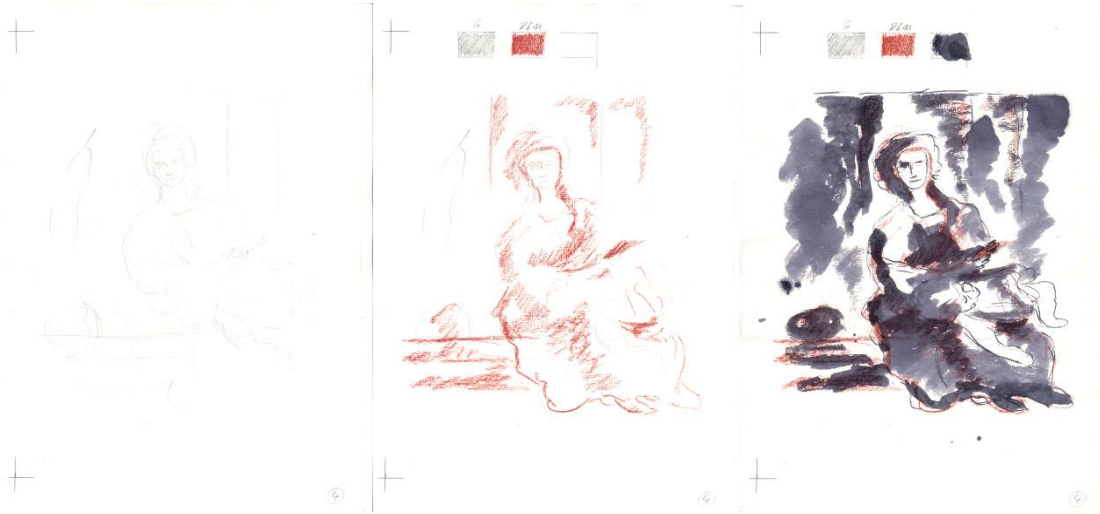

Figure 3: Sample layers from the data of the creation process as basis of evaluation. Left side: Step1 - First graphite sketch. Center: Step 2 - Underdrawings with red chalk. Right side: Steps 3 and 4 - Drawing with pen and iron gall ink plus final wash with two dilutions of ink in "two bowl technique", as described by Armenini (1587, pp.54-55). and Meder (1919, pp.54-55). Delineation after: Stefano della Bella, "Mother with two children", Florence, Galleria degli Uffici, Gabinetto Disegni e stampe, Inv.-Nr. 5937S

For imaging we used a multi-spectral camera equipped with a CMOS sensor capable of capturing the spectrum in a wavelength range of 400nm to 1000nm. Instead of subdividing the light into only three channels (red, green, blue = RGB), the camera divides it into 1040 different spectral bands. This allows us to differentiate materials of different physical compositions from their reflection patterns after processing the data with a cascade of computational algorithms.

**Workflow**

The workflow for analysing the drawings is shown in Figure 4 and explained below.

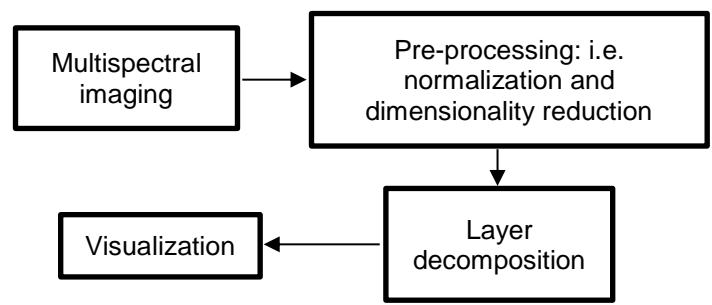

Figure 4: Proposed workflow

*Color Channel Normalization*

The sensitivity of the multispectral sensor varies across the captured spectrum. Different channels exhibit varying levels of brightness, which makes analysis of the images more difficult. To avoid these differences in brightness, we captured an image of a homogeneous white surface before scanning the drawing. In the scan of the drawing, each channel is corrected with a multiplicative factor, so that the sensor response on the white surface is homogeneous (Figure 5).

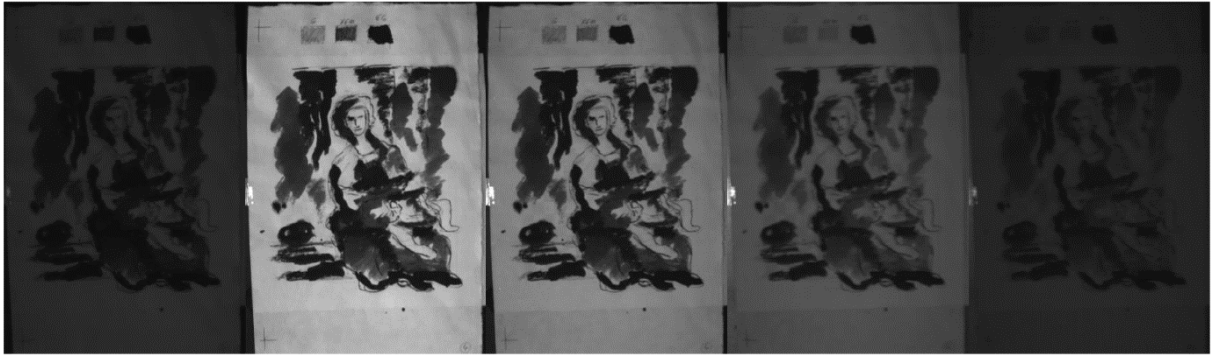
Figure 5(a)

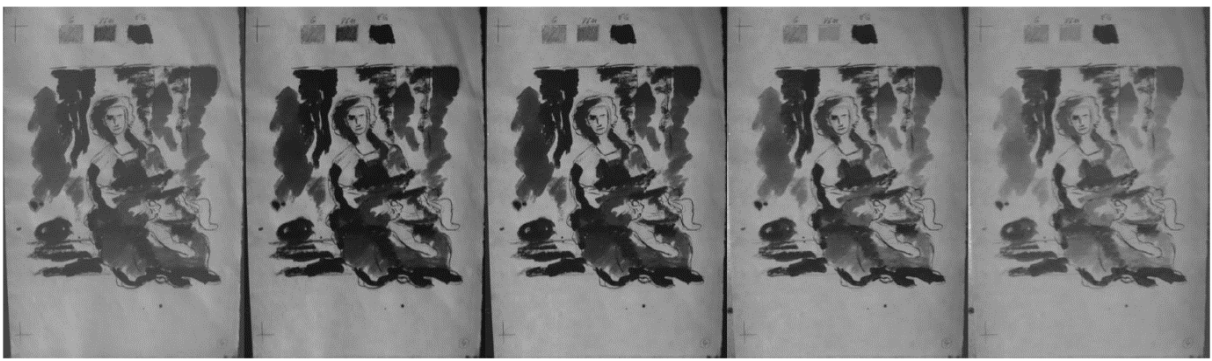
Figure 5(b)

Five sample spectral channels of a 258-band multispectral image. Images from left to right correspond to channels 31, 76, 100, 173 and 205 respectively. (a) Before normalization; (b) After normalization.

*Binning and Dimensionality Reduction on the Spectral Bands*

The hyperspectral camera we use operates at wavelengths of 400nm-1000nm. It is capable of capturing an image with 1040 spectral bans with a wavelength resolution of 0.58nm. However, images that are acquired with such a high wavelength resolution are typically very noisy. A common approach to reduce the noise is to average (bin) several spectral channels into one. In addition, the signal to noise ratio in the first and last spectral bands is very low. We therefore removed the first and last four spectral channels. Then we averaged every four neighbouring channels, i.e. binning was set to four. As a result, this process reduces the dimensionality to 258.

The 258 spectral bands make it challenging to analyse the multispectral image directly. In mitigation, a common approach in multispectral image processing is to greatly reduce the dimensionality before further processing (Harsanyi 1994). During dimensionality reduction, most of the spectral bands are either discarded (because they are found to be uninformative) or summarized using a linear or non-linear mapping function. Principle component analysis (PCA), introduced by Pearson (1901) and Hotelling (1933), is probably the most popular tool for this task (Jolliffe 2002, Pref. to 1.ed. ix; pp. 1-2; 78-80). We use PCA to summarize this information in only 5 channels (mathematically, this corresponds to preserving 99.5% of the image spectrum variance in our data). By using a highly reduced number of channels for image representation, the subsequent steps of processing also become more robust.

*Clustering*

Drawing regions of identical chalk or ink show a similar reflectance behaviour. To group image points of similar reflectance, we experimented with clustering algorithms. Specifically, we compared the performance of the K-means algorithm (Lloyd 1982) and Gaussian mixture models (GMM) (Friedman 1997). In both cases, we expected that each region of chalk or ink would be sorted into an individual cluster.

**Results and Evaluation**

To evaluate the data, we first computed the layer separation using multispectral data. Then, we compared the result of this computation to the individual (known) layers from the scanned images captured during the creative process. The resulting image is shown in Figure 6 (computed from the input data in Figure 5). On the left, the separation result of the K-means algorithm is shown; on the right, the result for GMM clustering. Identical grayscale values indicate identical cluster membership. K-means grouped the background into two clusters (most likely due to inhomogeneous illumination), while GMM isolated the background into a single cluster. It also turned out that GMM performed better in separating diluted inks, which can be best seen in the top right of the seated person in the picture. We tentatively conclude that GMM might be a better method for isolating sketch layers.

**Conclusion**

From our experiments, we present initial findings towards closing one of the diagnostic gaps in drawing analysis by using multispectral analysis and pattern recognition. We have shown that it is possible to extract the very specific wavelength signatures for red chalks when they are mixed with coloured overlaying ink. This occurs when the ink is semi-translucent, but also when it forms nearly opaque layers (Fig. 7). This suggests that our analysis can be applied to a broad range of material combinations found in drawings of various periods. This significantly expands the possible applications for such a tool beyond the scope of Old Master drawings to all forms of manuscripts.

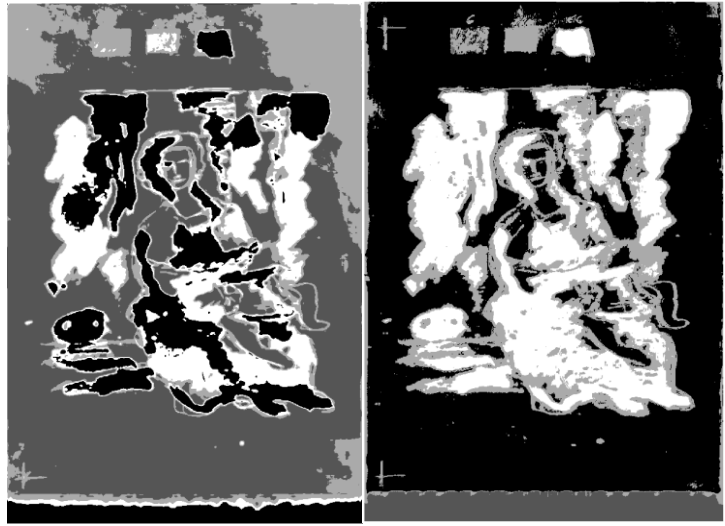

K-means result           GMM result

Figure 6: Clustering results

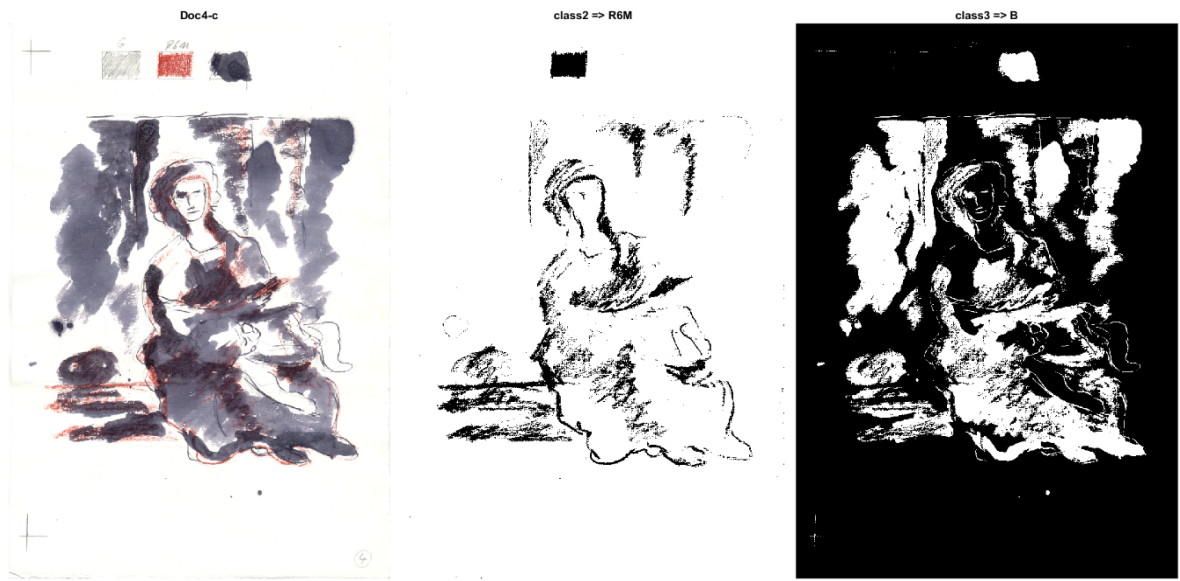

Figure 7: Layer separation of graphite and red chalk underdrawings overlaid with iron gall ink
Left: RGB-scan; Centre: Separation of the red chalk layer. The chalk could be displayed even when overlaid with a dense film of ink. The graphite underdrawings are not displayed here.
Right: Separation (inverse display) of the ink (pen drawing and wash)

From this starting point, we have expanded our study to larger datasets to evaluate quantitatively the performance of sketch layer separation. For quantitative evaluation, it is important to map the multispectral picture to the known layers pixel by pixel, which can be carried out by using an image registration algorithm. In a narrower sense, this improves the limits of the sensitivity and the resolving capacity of the technique.

While developing the procedure, we identified a smaller set of Old Master drawings from the seventeenth century, mainly by Nicolas Poussin (*1594 - †1665) and Stefano della Bella (*1610 - †1664), which we will examine in an forthcoming series of case studies. Furthermore, we would like to extend the experiments to address the additional problems exposed at the beginning of our analysis. It may be possible to distinguish the representations or writings on the verso of documents glued on a backing from the delineation of the recto with a series of similar algorithms in order to separate the distinct layers. That being said, the reconstruction of erasures, damages and surface abrasions will probably require even more refined or combined techniques.

This approach has the potential to capture the entire spectrum of drawing materials with one transportable camera. It also avoids radiation damage to the work of art under examination by finding specific set-ups of optimal wavelength channels to capture the raw image data in combination with optimized lightning spectra. These advantages could extend imaging-based art historical research to smaller museums or private collections, which are often not directly connected to centres for the technical investigation of artworks. Our analytic procedure provides a method that extends the range of approaches currently in use and could have broader applications for art history and the conservation of art.

**PICTURE CREDITS**
Fig. 1 Retrieved from http://smb-digital.de/eMuseumPlus?service=ExternalInterface&module=collection&objectId=1007411&viewType=detailView (23.02.2017), © Photo: Kupferstichkabinett der Staatlichen Museen zu Berlin - Preußischer Kulturbesitz, Photographer: Jörg P. Anders
Figs. 2-7 © A. Davari, A. Häberle, V. Christlein, A. Maier, C. Riess

**ACKNOWLEDGEMENTS**
The authors thank Tiffany Racco and Tiffany Hunt for their kind advice.